\pdfoutput=1
\documentclass[11pt]{article}
\usepackage[final]{acl}
\usepackage{times,latexsym}
\usepackage[T1]{fontenc}
\usepackage[utf8]{inputenc}
\usepackage{microtype}
\usepackage{inconsolata}
\usepackage{graphicx}
\usepackage{booktabs}
\usepackage{amsmath}
\usepackage{url}
\usepackage{tabularx}
\usepackage{booktabs}
\usepackage{hyperref}
\usepackage{booktabs}

\title{BD at BEA 2025 Shared Task: 
MPNet Ensembles for Pedagogical Mistake Identification and Localization in AI Tutor Responses}

\author{Shadman Rohan \quad Ishita Sur Apan \quad Muhtasim Ibteda Shochcho \quad Md Fahim \\
\textbf{Mohammad Ashfaq Ur Rahman \quad AKM Mahbubur Rahman \quad Amin Ahsan Ali} \\
Center for Computational \& Data Sciences, Independent University, Bangladesh (IUB) \\
\texttt{\{shadmanrohan, ishitasurapan, sho25100, fahimcse381, imashfaqfardin\}@gmail.com} \\
\texttt{\{akmmrahman, aminail\}@iub.edu.bd}}

\begin{document}
\maketitle

\begin{abstract}
We present Team BD's submission to the BEA 2025 Shared Task on Pedagogical Ability Assessment
of AI-powered Tutors, under Track 1 (Mistake Identification) and Track 2 (Mistake
Location). Both tracks involve three-class classification of tutor responses in educational dialogues –
determining if a tutor correctly recognizes a student’s mistake (Track 1) and whether the tutor pinpoints the
mistake’s location (Track 2). Our system is built on MPNet, a Transformer-based language model
that combines BERT and XLNet’s pre-training advantages. We fine-tuned MPNet on the task data using a
class-weighted cross-entropy loss to handle class imbalance, and leveraged grouped cross-validation (10
folds) to maximize the use of limited data while avoiding dialogue overlap
between training and validation. We then performed a hard-voting ensemble of the best models from
each fold, which improves robustness and generalization by combining multiple classifiers . Our
approach achieved strong results on both tracks, with exact-match macro-F1 scores of approximately 0.7110
for Mistake Identification and 0.5543 for Mistake Location on the official test set. We include comprehensive
analysis of our system’s performance, including confusion matrices and t-SNE visualizations to interpret
classifier behavior, as well as a taxonomy of common errors with examples. We hope our ensemble-based approach and findings provide useful insights for
designing reliable tutor response evaluation systems in educational dialogue settings.
\end{abstract}

\section{Introduction}
\label{sec:intro}

Effective intelligent tutoring systems need to be able to recognize and address student mistakes during interactions. To evaluate such capabilities in automated systems, the BEA 2025 Shared Task introduced a multi-dimensional assessment of AI tutor responses. In particular, Track 1 focuses on whether a tutor’s response identifies the student’s mistake, and Track 2 on whether it locates the mistake in the student’s answer. Each track is framed as a three-way classification: the tutor either fully recognizes/locates the error (“Yes”), partially or uncertainly does so (“To some extent”), or fails to do so (“No”). These pedagogically motivated categories draw from prior frameworks in educational dialogue analysis—for example, Mistake Identification corresponds to the student understanding dimension in Tack and Piech’s schema \cite{tack2022schema} and correctness in other tutoring evaluation schemata, reflecting how well the tutor acknowledges the student’s misconception.

Assessing tutor responses along such dimensions is challenging due to the nuanced and subjective nature of pedagogical feedback. Prior work has highlighted the lack of standardized evaluation criteria for dialogue quality. For instance, different studies have used varied measures (e.g., “speaking like a teacher,” “understanding the student,” etc.) to judge tutor responses. The BEA 2025 shared task addresses this gap by defining clear categories and metrics for evaluation \cite{bea2025sharedtask}. However, even with a fixed taxonomy, classifying responses correctly remains non-trivial: tutors may implicitly acknowledge an error without stating it outright, or they might hint at the error’s location in vague terms. Distinguishing between a definite “Yes” and a tentative “To some extent” thus requires subtle interpretation of language.

In this paper, we describe Team BD’s approach for automating annotation of mistake identification and mistake location, where we tackle the above challenges using a fine-tuned MPNet ensemble. MPNet is a pretrained Transformer that leverages masked and permuted language modeling to capture token dependencies, and it has demonstrated superior performance to earlier models like BERT, XLNet, and RoBERTa on language understanding benchmarks. We chose MPNet \cite{song2020mpnet} as our backbone to benefit from its strong generalization, while adapting it to this task through careful fine-tuning. Given the limited size of the labeled data (~2.5k examples), we employed grouped cross-validation to train multiple models without losing any training examples to a fixed held-out set. By grouping data by dialogue, we ensure that no examples from the same conversation appear in both training and validation, avoiding context leakage. We then ensemble the resulting models via majority voting, a strategy known to reduce variance and improve robustness by combining diverse classifiers. Ensemble learning has a rich history of boosting performance in machine learning and is particularly useful when individual models can capture complementary aspects of the data or errors. Our main contributions are:
\begin{itemize}
    \item We develop an ensemble-based system for tutor response classification, fine-tuning MPNet with cost-sensitive learning to handle class imbalance.
    
    \item Our approach achieves strong accuracy and macro-F1 scores across both tracks, demonstrating the effectiveness of cross-validation ensembling.
    
    \item We conduct detailed error analysis, including latent space visualization and class confusion patterns, revealing systematic issues such as confusion between full and partial mistake recognition.
    
    \item We discuss key limitations—such as calibration, category subjectivity, and model constraints—and suggest future directions like leveraging larger language models and multi-task learning.
\end{itemize}

\section{Related Work}
\label{sec:related}
\textbf{Evaluation of Tutor Responses}: The task of judging tutor or teacher responses in educational dialogues has recently garnered attention. Tack and Piech \cite{tack2022ai} introduced the AI Teacher Test to measure the pedagogical ability of dialogue agents, proposing dimensions such as whether the agent understands the student’s error and provides helpful guidance. This laid groundwork for defining concrete evaluation criteria. Following this, the BEA 2023 Shared Task  \cite{tack2023bea} focused on generating AI teacher responses (rather than classification), where models like GPT-3 and Blender were challenged to produce tutor-like feedback. That work underscored the need for effective evaluation metrics to assess the quality of generated responses. The BEA 2025 Shared Task \cite{bea2025sharedtask} moves a step further by creating a benchmark dataset of tutor responses annotated along multiple pedagogical dimensions. The dataset leverages dialogues from \textbf{MathDial} \cite{macina2023mathdial} and \textbf{Bridge} \cite{wang2024bridge}, two collections of student-tutor interactions in the math domain. Each tutor response in these dialogues was labeled by experts as to whether it identifies the student’s mistake, pinpoints the mistake’s location, provides guidance, and offers actionable next steps. Such multi-faceted annotation of tutor feedback is relatively novel; it connects to earlier work on dialogue act classification \cite{stolcke2000dialogue} in that both involve categorizing utterances, but here the labels are pedagogical quality ratings rather than communicative intent.

\textbf{Ensemble Methods in NLP}: Ensemble learning has long been used to improve predictive performance by combining multiple models. Classic studies, such as Dietterich's work on ensemble methods, demonstrated that an ensemble of diverse classifiers can correct individual models’ errors and reduce variance \cite{dietterich2000ensemble}. In NLP tasks, ensembling fine-tuned language models has proven effective for boosting accuracy and robustness. For instance, \cite{ovadia2019can} and \cite{gustafsson2020evaluating} found that deep ensembles improve reliability under dataset shift. In shared task competitions, top teams often resort to model ensembling to squeeze out additional performance—the combined vote of models tends to outperform any single model alone. These benefits come at the cost of increased computational overhead, but given our task’s moderate scale, an ensemble of MPNet classifiers was feasible. Our approach aligns with this trend, as we build an ensemble of 10 MPNet-based classifiers (from cross-validation folds) to tackle the classification of tutor responses.

\textbf{Dialogue and Educational NLP}: Related to our work is research on grammatical error detection and correction, where systems identify mistakes in student-written text. Notably,  \cite{ng2014conll} and  \cite{bryant2019bea} have contributed significantly to this field. However, our task differs in that the “mistakes” are conceptual or procedural errors in a problem solution, and we are evaluating the tutor’s response to those errors rather than directly analyzing the student’s text. Another line of relevant work is on student response analysis in tutoring systems, where the goal is to classify student answers as correct, incorrect, or incomplete.  \cite{dzikovska2013semeval} explored this in the context of the SemEval-2013 Task 7. In our case, the roles are reversed—we classify the tutor’s replies. Nonetheless, techniques such as using pre-trained language models and handling class imbalance are common challenges across these domains. We also draw on insights from educational dialogue analysis: studies like \cite{daheim2024stepwise} examined tutor responses for targetedness and actionability, which correspond to our Track 2 and Track 4 tasks. These studies emphasize the subtle linguistic cues that indicate whether a tutor has pinpointed an error (e.g., referencing a specific step in the student’s solution) or just given generic feedback.

In summary, our work is situated at the intersection of dialogue evaluation and text classification. We build upon the shared task’s provided taxonomy \cite{bea2025sharedtask} and prior educational NLP research, employing modern Transformer models and ensemble techniques known to be effective in such tasks.

\section{Data and Task Definition}
\label{sec:data}
\textbf{Task Definition:} Tracks 1 and 2 are classification tasks applied to tutor responses in a dialogue. Based on the previous conversation history between students and tutors, in Track 1
(Mistake Identification), the system must determine if the tutor’s response indicates recognition of the
student’s mistake. In Track 2 (Mistake Location), the system judges if the tutor points out the specific location
or nature of the mistake in the student’s solution. Both tasks share the same label set: \textbf{Yes, To some
extent,} or \textbf{No}. Because these categories can be nuanced, the shared task also defined a lenient evaluation where “Yes”
and “To some extent” are merged, but our system is trained on the full 3-class distinction (exact
evaluation).

\textbf{Dataset}: The training (development) data provided by the organizers consists of annotated educational
dialogues in mathematics, drawn from the MathDial and Bridge datasets. Each dialogue includes a
student’s attempt at a math problem (containing a mistake or confusion) and one or more tutor responses (from either human tutors or various LLMs such as Mistral, Llama, GPT-4, etc. acting as tutors). Each tutor response is
annotated with the three-class labels for all four dimensions (Tracks 1–4). In total, the development
set contains 300 conversation history and over 2,480 tutor responses with annotations. On average, each
dialogue context yields 8–9 different tutor responses (one from each of several tutor sources), which were
all annotated. The test set is constructed in the same way but uses held-out dialogues and responses—both the ground-truth labels and the tutors’ identities are hidden.


The development set for both \textbf{Track 1 (Mistake Identification)} and \textbf{Track 2 (Mistake Location)} consists of the same 300 dialogues and 2,476 tutor responses. However, the label distributions differ between tracks due to the nature of the classification tasks. The underrepresentation of the \textit{To some extent} class in both tracks poses challenges for model learning.

\begin{table}[t]
\centering
\small
\begin{tabular}{l c}
\toprule
\textbf{Model} & \textbf{Macro-F1 Score} \\
\midrule
BERT-large & 0.6851 \\
DeBERTa    & 0.6845 \\
MPNet (selected) & \textbf{0.6975} \\
\bottomrule
\end{tabular}
\caption{10 fold Cross-validation Macro-F1 scores for different Transformer models on the track 1 development set. MPNet achieves the highest score.}
\label{tab:model-comparison}
\end{table}

\section{Methodology}

\subsection{Preprocessing}

All tutor responses and conversation histories were first lowercased (while preserving punctuation) to ensure consistent casing.

To standardize and sanitize the responses, we applied a series of targeted cleaning steps:

\begin{itemize}
    \item \textbf{Extra Info Removal}: Eliminated any metadata or annotations not part of the tutor’s actual reply.
    \item \textbf{Appended Dialogue Trimming}: Removed follow-up conversational turns that were appended after the original tutor response (e.g., speculative follow-up questions or acknowledgments).
    \item \textbf{Code Abstraction}: Replaced Python code blocks with the placeholder \texttt{<<python code>>} to retain structural intent while abstracting away executable details.
    \item \textbf{Punctuation Cleanup}: Stripped redundant or mismatched punctuation (e.g., extraneous quotes or dashes) that might confuse the tokenizer or the model.
\end{itemize}

\autoref{tab:preprocessing-wide} provides a summary of how many instances were affected by each category. We observed that models such as \textbf{Phi-3} and \textbf{Llama-3.1-405B} required the most extensive preprocessing.

Finally, each input example—consisting of the conversation history, cleaned response, and separator tokens—was constrained to a maximum of 512 MPNet tokens. In cases where the input exceeded this limit, we manually pruned low-value content (e.g., greetings or small talk) from the conversation history to retain the most relevant context.

\subsection{LM Finetuning}
In our experiments, we utilize transformer-based pretrained language models (LMs). 
Since these models may lack task-specific contextual knowledge, we fine-tune them 
on our target tasks to improve performance.

To begin, we consider a pretrained language model denoted as $\phi_{\text{LM}}$. 
Each tutor's response after preprocessing $T$ is input to the model, yielding a sequence of tokens 
$T = \{ t_{[\text{CLS}]}, t_1, t_2, \ldots, t_n \}$ along with their corresponding 
layer-wise hidden representations $H^l = \{ h_{[\text{CLS}]}^l, h_1^l, h_2^l, \ldots, h_n^l \}$.

In our setup, we use the hidden representation of the [CLS] token from the final 
layer as the sentence-level representation of the input $T$, defined as:
\[
h_T = \phi_{\text{LM}}(T)^L_{[\text{CLS}]} = H^L_{[\text{CLS}]}
\]

This representation $h_T$ is then passed through a classification head to produce 
the prediction. The classification head consists of a dropout layer $\textit{Drop}$ followed 
by a linear transformation:
\[
p = W \cdot \text{Drop}(h_T) + b
\]

Finally, we use a cross-entropy loss function to update the parameters of the 
language model $\phi_{\text{LM}}$ during training.

\subsection{Grouped Cross-Validation}

We employ \textit{group cross-validation} to ensure robust evaluation and mitigate overfitting. 
In this approach, each dialogue (or group of dialogues) is entirely assigned to either 
the training or validation set within each fold, preventing shared context between the 
training and validation sets.

For each fold $f \in \{1, 2, \ldots, k\}$, we define the training and validation sets 
as $\mathcal{G}^{(f)}_{\text{train}}$ and $\mathcal{G}^{(f)}_{\text{val}}$, respectively, 
where each set contains whole dialogues (or groups) with no overlap.
We monitor the model's performance on the validation set using the \textit{macro-averaged F1 score}
(macro-F1), which provides a balanced measure of performance across classes. For each fold, 
we save the model checkpoint that achieves the highest macro-F1 score on the validation set.

The final performance of the model is computed by aggregating the macro-F1 scores 
across all $k$ folds.

\subsection{Ensembling Strategy}
To enhance model performance, we employed an \textit{ensembling strategy} where the top-performing models from each fold were combined using hard voting. Specifically, for each track, we had a total of $N = 10$ models (one from each fold).

Let $\hat{y}^{(f)}_i$ denote the prediction of the model from fold $f$ for the $i$-th sample, where $f \in \{1, 2, \ldots, N\}$. The final prediction $\hat{y}_i$ for each sample $i$ was determined by majority vote:

\[
\hat{y}_i = \text{mode}(\hat{y}^{(1)}_i, \hat{y}^{(2)}_i, \ldots, \hat{y}^{(N)}_i)
\]

In the case of a tie, the tie-breaking rule was based on the average softmax confidence across all models. Let $s^{(f)}_i$ denote the softmax output (confidence) of the $f$-th model for the $i$-th sample. If a tie occurs, the final prediction is chosen as:

\[
\hat{y}_i = \arg\max\left( \frac{1}{N} \sum_{f=1}^{N} s^{(f)}_i \right)
\]

Ensembling helps to reduce variance and correct individual model biases, leading to more robust predictions. Our ensembling approach improved the macro-F1 score by 2–3 points over the performance of individual models.


\section{Experimental Setup}

\subsection{Implementation Details}

\noindent \textbf{Model Selection} In our experiments, we compared several such models—including BERT-large, DeBERTa, and MPNet—on a held-out subset of the training data. Among these, MPNet achieved the best macro-F1 score (see Table~\ref{tab:model-comparison}), and was thus selected as our backbone.

\noindent \textbf{Software and Package} We conducted all experiments using Python~3.9, PyTorch~1.13, and the Hugging Face Transformers library (v4.x) \cite{wolf-etal-2020-transformers}. Specifically, we fine-tuned the \texttt{sentence-transformers/all-mpnet-base-v2} model available on the Hugging Face Model Hub \cite{all-mpnet-base-v2}. Tokenization was performed using MPNet's tokenizer, with inputs truncated to a maximum length of 300 tokens.

The models were trained on a single NVIDIA Tesla V100 GPU (16\,GB). Each fold took approximately 3--5 minutes per epoch to train, with convergence typically reached within 3 epochs (i.e., 10--15 minutes per model). Full ensemble training (10 models for Track~1 and 7 for Track~2) required a few hours in total. Despite the ensemble size, inference was efficient: classifying the entire test set (several hundred responses) took under a minute.

\noindent \textbf{Model Hyperparameters}

We used the AdamW optimizer with a learning rate of \(2 \times 10^{-5}\), selected through preliminary experiments on a held-out validation set. This setting outperformed alternative learning rates such as \(1 \times 10^{-5}\) and \(3 \times 10^{-5}\) in terms of macro-F1. A linear learning rate decay schedule was used, along with early stopping based on validation macro-F1 (patience = 2 epochs). We trained with a batch size of 32 and applied a dropout rate of 0.1 in the classification head. No gradient accumulation was used.

\noindent \textbf{Handling Class Imbalance}
To address class imbalance and improve robustness, we experimented with minor data augmentation by substituting alternate tutor responses in training examples; however, this yielded minimal benefit. Evaluation metrics and confusion matrices were computed using Scikit-learn. For reproducibility, we set fixed random seeds and enabled deterministic behavior where feasible.






To mitigate class imbalance, we used a class-weighted cross-entropy loss, where the weight for each class \( c \) was computed as:
\[
w_c = \frac{N}{K \cdot n_c}
\]
with \( N \) being the total number of samples, \( K \) the number of classes, and \( n_c \) the count for class \( c \). This formulation emphasizes underrepresented classes without overly penalizing frequent ones.

For Track~1 (Mistake Identification), class distributions were skewed toward ``Yes'' (1932), compared to ``No'' (370) and ``To some extent'' (174). We thus used the weight vector:
\[
[w_{\text{No}}, w_{\text{Some}}, w_{\text{Yes}}] = [1.0, 3.0, 0.5]
\]
to boost recall for the rare ``Some extent'' class and mildly down-weight the majority class.

In Track~2 (Mistake Location), the frequencies were: ``Yes'' (1504), ``No'' (732), and ``To some extent'' (240). Based on this, we used:
\[
[w_{\text{No}}, w_{\text{Some}}, w_{\text{Yes}}] = [0.8, 2.2, 0.9]
\]
These weights, derived from inverse class frequencies and lightly tuned, improved macro-F1 by reducing systematic underprediction of minority classes. Although not extensively optimized, this approach provided consistent performance gains across both tracks.

\subsection{Evaluation Metrics}
Following the shared task guidelines, we report both Accuracy and Macro F1. Macro F1, the unweighted average of per-class F1 scores, is emphasized due to class imbalance. We monitored performance using these metrics on the validation set during training and evaluated on the aggregated development set using cross-validation predictions. Final test metrics were provided by the organizers. We focus on exact 3-class classification; lenient 2-class metrics (merging ``Yes'' with ``To some extent'') were higher but are omitted here for brevity.

\section{Result and Analysis}

\begin{table}[t]
\centering
\small
\begin{tabular}{@{}lcc@{}}
\toprule
\textbf{Track} & \textbf{Macro F1} & \textbf{Accuracy} \\
\midrule
\multicolumn{3}{@{}l}{\textit{Track 1 – Mistake Identification}} \\
Best (BJTU)           & 0.718 & 0.862 \\
Ours (Test)           & 0.711 & 0.877 \\
Ours (CV aggregate)   & 0.685 & 0.869 \\
\midrule
\multicolumn{3}{@{}l}{\textit{Track 2 – Mistake Location}} \\
Best (BLCU-ICALL)     & 0.598 & 0.768 \\
Ours (Test)           & 0.554 & 0.714 \\
Ours (CV aggregate)   & 0.560 & 0.700 \\
\bottomrule
\end{tabular}
\caption{Comparison of our system’s macro-F1 and accuracy with top leaderboard scores on both tracks.}
\label{tab:results-comparison}
\end{table}

\subsection{Main Result}
To contextualize our system’s performance, we compared it against the top submissions from the official shared task leaderboard. On Track1 (Mistake Identification), our model achieved a macro-F1 of 0.711 on the test set, placing 5th out of 44 participating teams. The top-ranked system (BJTU) achieved a macro-F1 of 0.718, indicating that our system performs competitively, within 0.7 points of the best result. For Track2 (Mistake Location), our system scored 0.554 macro-F1 on the test set, ranking 7th out of 31 teams. The highest score on this track was 0.598, obtained by BLCU-ICALL. While our model trails behind the top result by approximately 4.4 points in macro-F1, it still exceeds the median leaderboard performance. 

Our system achieved higher accuracy than the top Track~1 system (0.877 vs. 0.862), suggesting stronger performance on dominant classes, albeit with slightly lower balance across all classes.

Even though our system performs well, a closer examination of its errors provides insights into its decision-making and the task’s inherent difficulty. We carried out an error analysis on the development set predictions, focusing on confusion patterns and the nature of misclassified cases.

\subsection{Class-Level Performance Analysis}

\begin{figure}[ht]
\centering
\includegraphics[width=\columnwidth]{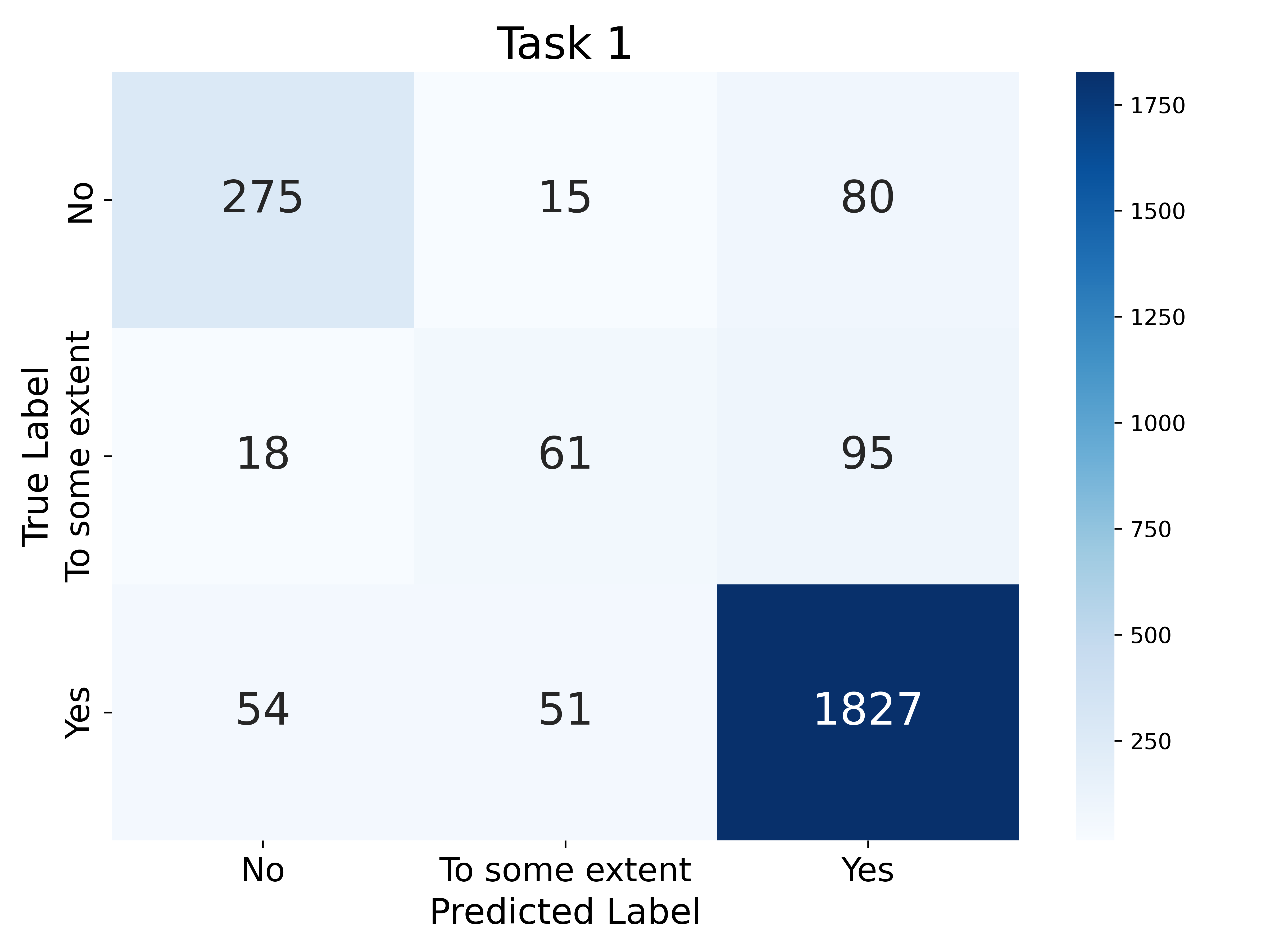}
\caption{Confusion matrix for Track 1 (Mistake Identification) on the development set. The model shows strong performance on the "Yes" class but has difficulty distinguishing partial acknowledgment ("To some extent").}
\label{fig:t1_confusion_matrix}
\end{figure}

\begin{figure}[ht]
\centering
\includegraphics[width=\columnwidth]{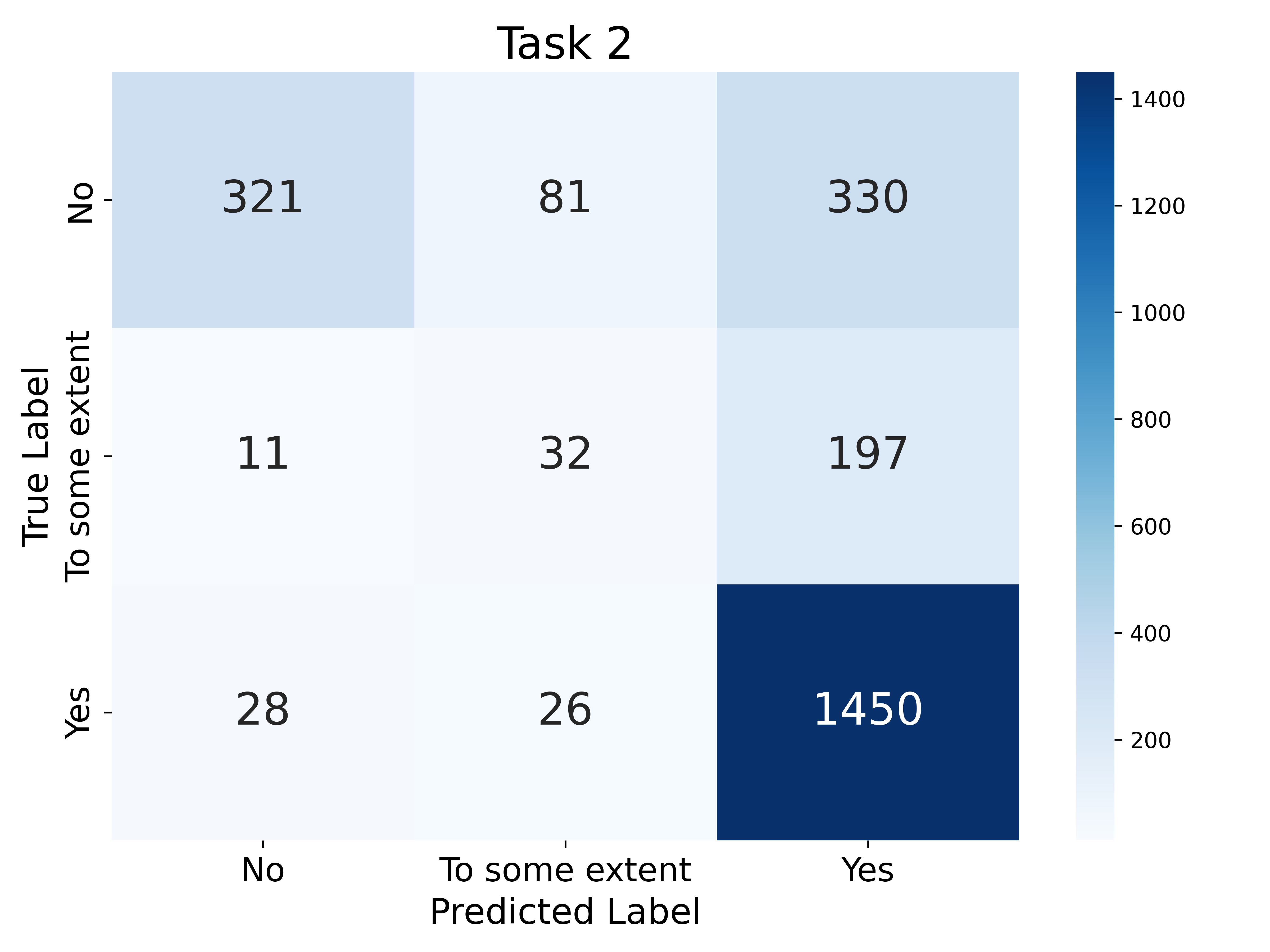}
\caption{Confusion matrix for Track 2 (Mistake Location). The model maintains high accuracy on explicit localizations ("Yes") but misclassifies many “To some extent” and “No” cases, highlighting the subtlety of location inference.}
\label{fig:t2_confusion_matrix}
\end{figure}

To gain insight into how well our system distinguishes among the three pedagogical feedback categories, we analyze confusion matrices for both tasks. Figures~\ref{fig:t1_confusion_matrix} and \ref{fig:t2_confusion_matrix} visualize model predictions against gold labels on the development set for Track 1 (Mistake Identification) and Track 2 (Mistake Location), respectively.

In Track 1 (Figure~\ref{fig:t1_confusion_matrix}), the model performs strongly on the "Yes" class, correctly identifying 1,827 instances, with relatively low misclassification into the "No" (54) and "To some extent" (51) classes. The "No" class is also well captured with 275 correct predictions and few false positives. The model struggles more with the "To some extent" category: 61 were correctly predicted, but 113 were misclassified as either "No" or "Yes." This aligns with our earlier claim that “To some extent” lies on a subjective continuum and is more difficult to pin down categorically.

For Track 2 (Figure~\ref{fig:t2_confusion_matrix}), a similar trend emerges. The model again shows high accuracy on “Yes” (1,450 correct), but struggles to distinguish “To some extent,” which is often misclassified as “Yes” (197 cases) or “No” (11 cases). Notably, the “No” class is less cleanly separated in Track 2 compared to Track 1, with 330 examples misclassified as “Yes.” This may suggest that tutors sometimes appear to reference an error without pinpointing its location, confusing the model’s judgment.

Overall, these confusion matrices illustrate the asymmetric difficulty across classes. "Yes" responses are most reliably predicted due to their clearer, more direct language. "To some extent" predictions remain a challenge, particularly when tutors use indirect or hedging phrasing that blurs the line between partial and full error acknowledgment or localization.

\subsection{Embedding Space Insights}

\begin{figure}[h]
\centering
\includegraphics[width=\columnwidth]{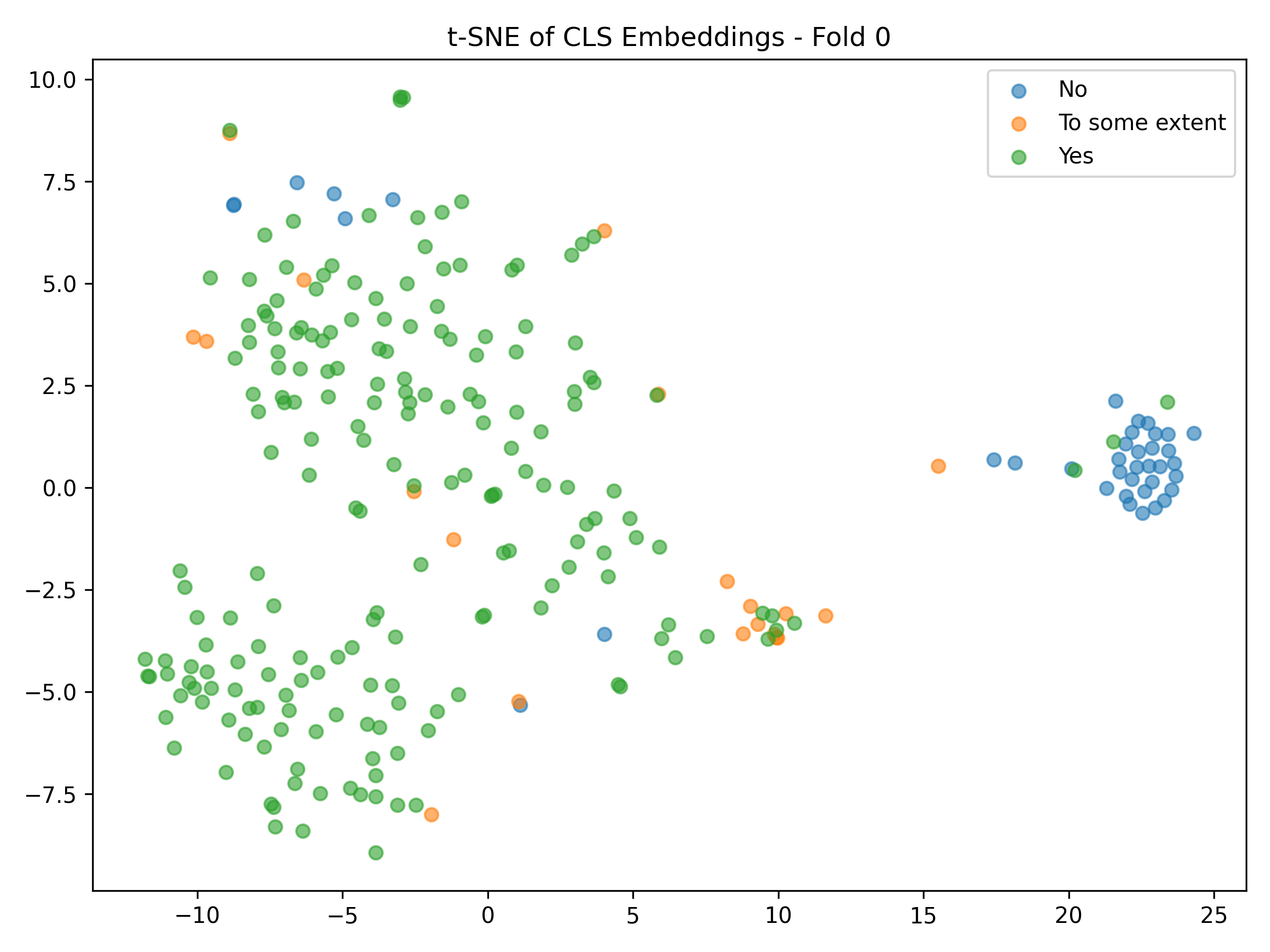}
\caption{t-SNE projection of [CLS] embeddings from the held-out fold (Fold 0), colored by true label. “Yes” and “To some extent” examples are scattered and intermixed, whereas “No” forms a more compact cluster, indicating lower intra-class variation.}
\label{fig:fold0_tsne_visualization}
\end{figure}

\begin{figure}[h]
\centering
\includegraphics[width=\columnwidth]{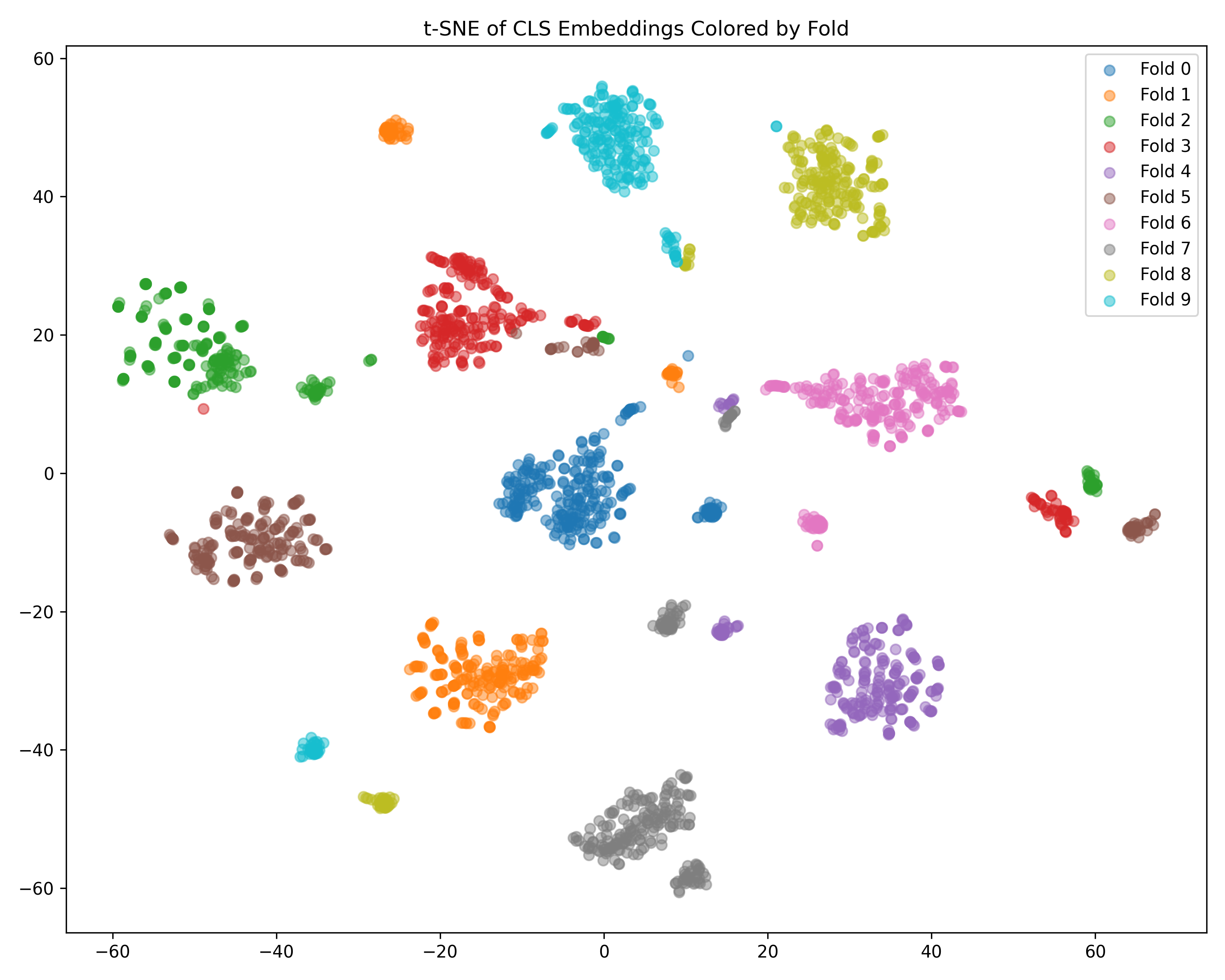}
\caption{t-SNE projection of [CLS] embeddings from MPNet models across all 10 cross-validation folds. Each point represents a tutor response from a held-out fold, colored by fold ID. The emergence of distinct clusters suggests that each fold-specific model learns a consistent but fold-specific embedding subspace, reflecting representational diversity across the ensemble.}
\label{fig:all_tsne_visualization}
\end{figure}

To better understand the internal representations learned by our model, we applied t-SNE \cite{vanDerMaaten2008} to the [CLS] embeddings from the final Transformer layer. These projections reveal how the model organizes tutor responses in the embedding space across folds and classes.

Figure~\ref{fig:all_tsne_visualization} shows the t-SNE projection of the [CLS] embeddings across all ten cross-validation folds, with points colored by fold ID. We observe that embeddings from each fold tend to form compact, well-separated clusters. This indicates that while training on different subsets, each fold-specific model learns fold-consistent but distinct representations. The tightness of these clusters also suggests good embedding stability and coherence across training runs.

Figure~\ref{fig:fold0_tsne_visualization} presents the t-SNE visualization for the held-out fold (Fold 0), this time colored by the true label. Unlike the per-fold visualization, class-level structure is less distinct: the “Yes” and “To some extent” responses are widely dispersed and often intermingle, suggesting overlapping semantic characteristics. In contrast, the “No” class forms a more compact group, indicating that tutor responses with no recognition of error share more consistent linguistic patterns. This aligns with our earlier findings that “Yes” and “To some extent” are harder to separate, as they exist on a continuum of acknowledgment.

Together, these visualizations support our earlier confusion matrix results and highlight the challenge of distinguishing nuanced pedagogical feedback categories based solely on language.

\subsection{Error Taxonomy}

To better understand where the model fails, we analyzed misclassified responses from both tasks and developed a taxonomy of recurring error types, summarized in Table~\ref{tab:error-taxonomy}. These categories reflect systematic issues in how the model interprets pedagogical language.

\begin{table*}[t]
\centering
\small
\begin{tabular}{p{3cm} p{5cm} p{7cm}}
\toprule
\textbf{Error Type} & \textbf{Description} & \textbf{Example Scenario} \\
\midrule
False Negative (Missed Signal) & Tutor indicates or locates a mistake, but the model predicts “No” or “To some extent.” & \textit{Tutor:} ``Can you check the multiplication again?'' \newline Gold: Yes $\rightarrow$ Pred: To some extent \\
False Positive (Over-interpretation) & Model predicts “Yes” despite the tutor giving no error feedback. & \textit{Tutor:} ``Let's try another one.'' \newline Gold: No $\rightarrow$ Pred: Yes \\
Partial–Full Confusion & Confuses indirect hints as full identification, or subtle localization as partial. & \textit{Tutor:} ``You're close, just verify your subtraction.'' \newline Gold: To some extent $\rightarrow$ Pred: Yes \\
Hedged Language Confusion & Tutor’s suggestion is misread due to polite phrasing or indirect cues. & \textit{Tutor:} ``Maybe revisit the earlier step?'' \newline Gold: Yes $\rightarrow$ Pred: To some extent \\
Contextual Miss & Misclassification caused by ignoring or misusing multi-turn context. & \textit{Tutor:} Feedback depends on an earlier step, but the model misses the reference. \\
Template Bias & Model favors phrases resembling training-time patterns, even when semantically incorrect. & \textit{Tutor:} ``Great work!'' with no correction. \newline Model assumes this implies error recognition. \\
\bottomrule
\end{tabular}
\caption{Taxonomy of common misclassification errors in both tasks, with representative examples.}
\label{tab:error-taxonomy}
\end{table*}

\textbf{False Negatives (Missed Signal).} These errors occur when the model fails to recognize that the tutor has identified or located a mistake, typically labeling the response as ``No'' or ``To some extent'' instead of ``Yes.'' Such cases often involve subtle cues like rhetorical questions or light correction phrasing (e.g., ``Can you check the multiplication again?''), which the model may under-interpret.

\textbf{False Positives (Over-interpretation).} Here, the model predicts ``Yes'' even when the tutor does not provide evidence of error recognition. This often results from over-interpreting generic encouragement (e.g., ``Let's try another one.'') or positive sentiment as pedagogical feedback.

\textbf{Partial–Full Confusion.} A frequent source of confusion is the distinction between full and partial identification or localization. Indirect language such as ``You're close, just verify your subtraction'' may be intended as partial feedback, but the model may treat it as a complete identification.

\textbf{Hedged Language Confusion.} Tutors often use polite or indirect language (e.g., ``Maybe revisit the earlier step?''), especially in educational settings. Such hedging may obscure intent, leading the model to underestimate the strength of the feedback signal.

\textbf{Contextual Miss.} Some misclassifications stem from failing to use conversational history. For instance, if a tutor's comment refers to an earlier incorrect step, the model may mislabel it when that context is not incorporated effectively.

\textbf{Template Bias.} We also observed that the model sometimes over-relies on surface patterns seen during training. For example, statements like ``Great work!'' may be incorrectly classified as ``Yes'' due to template bias, even when no mistake is acknowledged.

These error categories offer valuable insight into the linguistic and contextual challenges of the task. They suggest that improvements in discourse modeling, uncertainty handling, and pragmatic language understanding could further enhance performance.

From the above taxonomy, we see that many of the model’s mistakes correspond to understandable difficulties. False negatives often involved indirect tutor feedback—the tutor recognized the mistake but phrased it as a question or hint, requiring inference to identify it as an acknowledgment of error. Our model sometimes took such tentative language at face value and labeled it as if the tutor did nothing. False positives, on the other hand, were cases where the tutor’s response had reassuring or neutral language that the model mistook for a sign of recognizing a mistake. For example, tutors might say “Let’s double-check that” even when the student was correct (encouraging the student, not pointing an error), and the model erroneously flagged it as identifying an error.

The partial vs. full confusion category was the most prevalent error type. This reflects the inherent ambiguity of the “To some extent” class—even human annotators might differ on these in some cases. Our model would sometimes collapse it into one of the binary decisions (“Yes” or “No”) depending on slight wording differences. In some cases, the model predicted “To some extent” when the tutor had actually pinpointed the error but perhaps in a subtle way; in others, it predicted “Yes” for a tutor response that was only hinting. This suggests that improving the model’s understanding of nuanced language (perhaps via better context usage or training on more examples of hedging) could help.

We also found that ambiguous wording and polite phrasing (common in educational settings) posed challenges. Phrases like “Maybe check that again” require contextual understanding—they might indicate an error without explicit wording. Our model did catch many of these, but not all. Some errors could be attributed to the model’s lack of world knowledge or reasoning; for example, if a tutor says “Remember the formula for area,” the model needs to infer that the student likely made a mistake related to area calculation and that the tutor is hinting at it—a level of reasoning beyond surface text.

In summary, the error analysis reveals that while our ensemble is effective, there is room for improvement in handling borderline cases and understanding implicit signals. These findings guided us in considering potential enhancements, as discussed next.

\subsection{Confidence Distribution and Calibration}

\begin{figure}[t]
\centering
\includegraphics[width=\columnwidth]{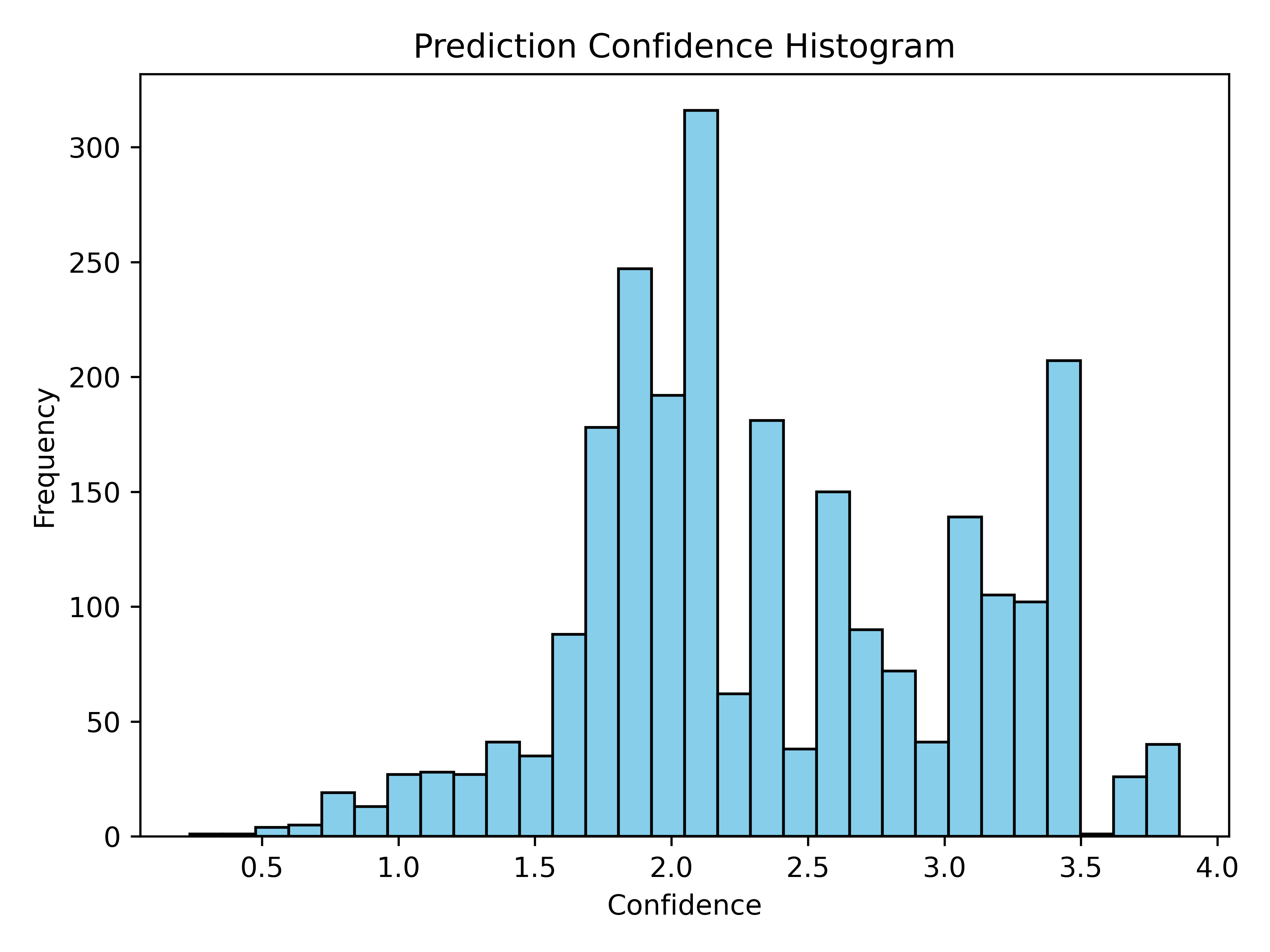}
\caption{Histogram of prediction confidence values for Track~1 (Mistake Identification). Most predictions fall within a mid-confidence range.}
\label{fig:task1_conf_hist}
\end{figure}

\begin{figure}[t]
\centering
\includegraphics[width=\columnwidth]{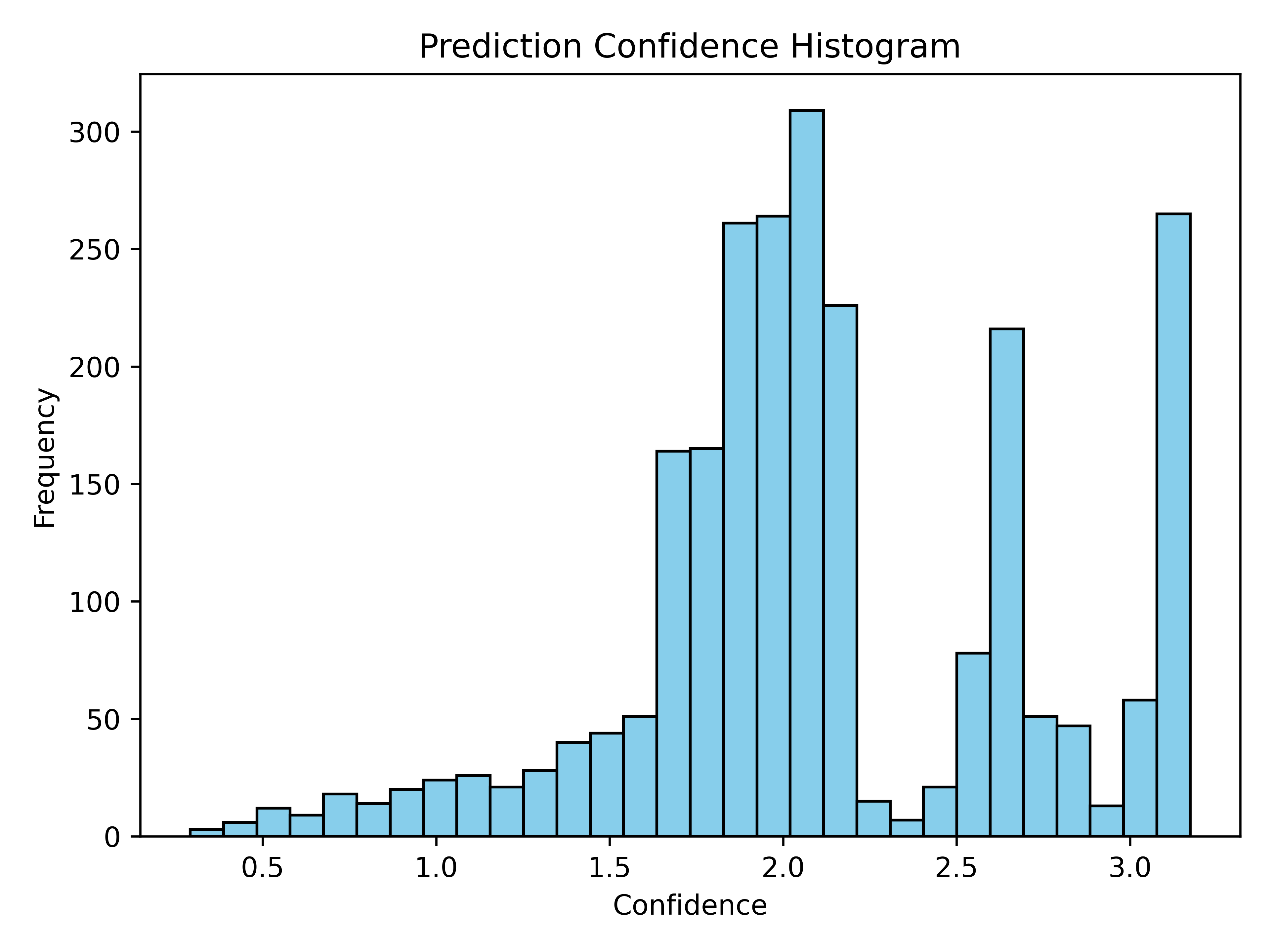}
\caption{Histogram of prediction confidence values for Track~2 (Mistake Location). A similar mid-range clustering pattern is observed, with some extreme confidence peaks.}
\label{fig:task2_conf_hist}
\end{figure}

To further investigate the model's decision-making behavior, we analyzed its prediction confidence across classes and tasks. Figures~\ref{fig:task1_conf_hist} and \ref{fig:task2_conf_hist} present histograms of predicted confidence scores for Track~1 and Track~2, respectively. These reflect the model's certainty in its predictions across the development set.

\begin{figure}[t]
\centering
\includegraphics[width=\columnwidth]{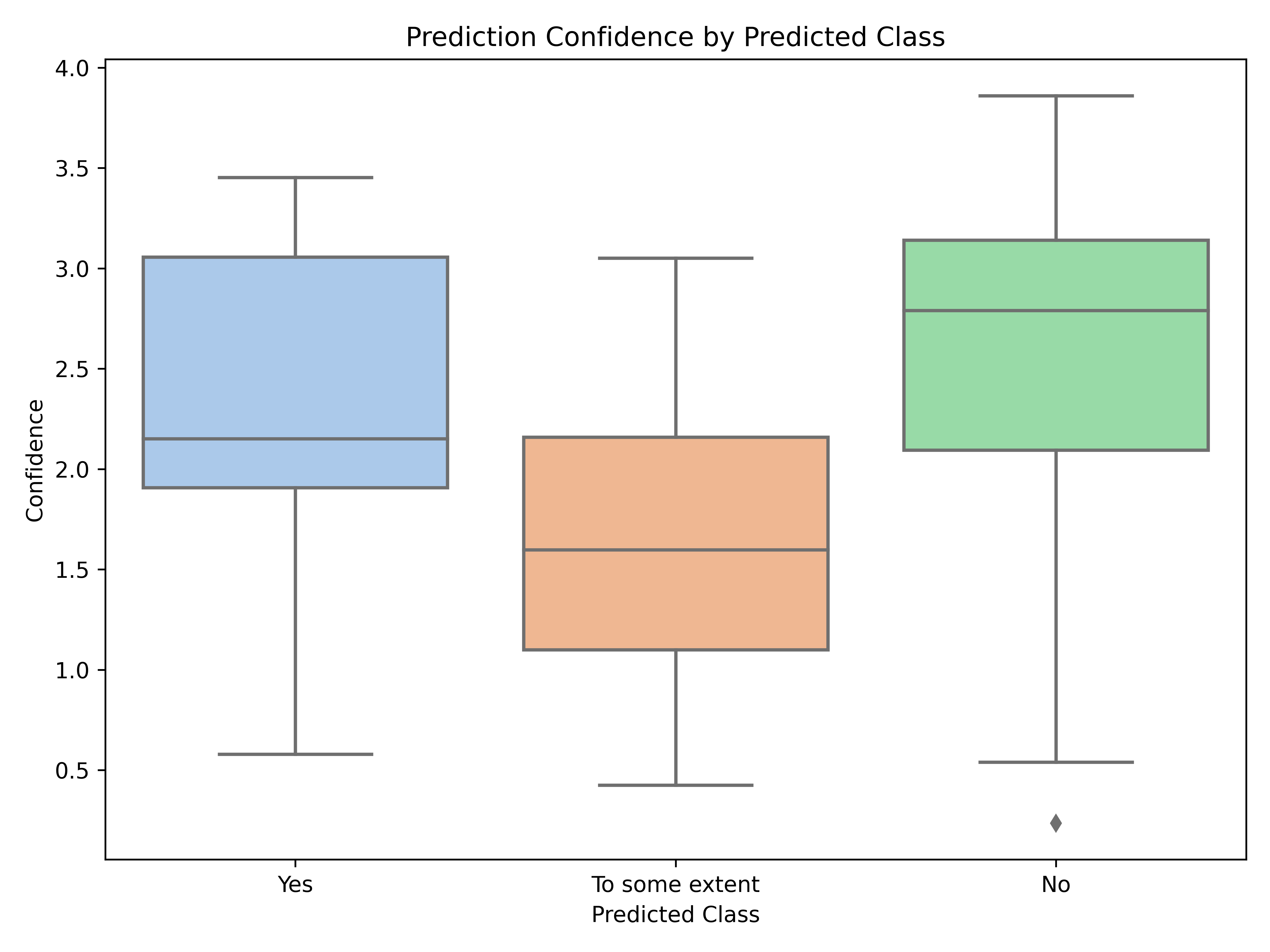}
\caption{Boxplot of confidence by predicted class (Track~1). Predictions labeled “To some extent” tend to have lower median confidence.}
\label{fig:task1_conf_box}
\end{figure}

\begin{figure}[t]
\centering
\includegraphics[width=\columnwidth]{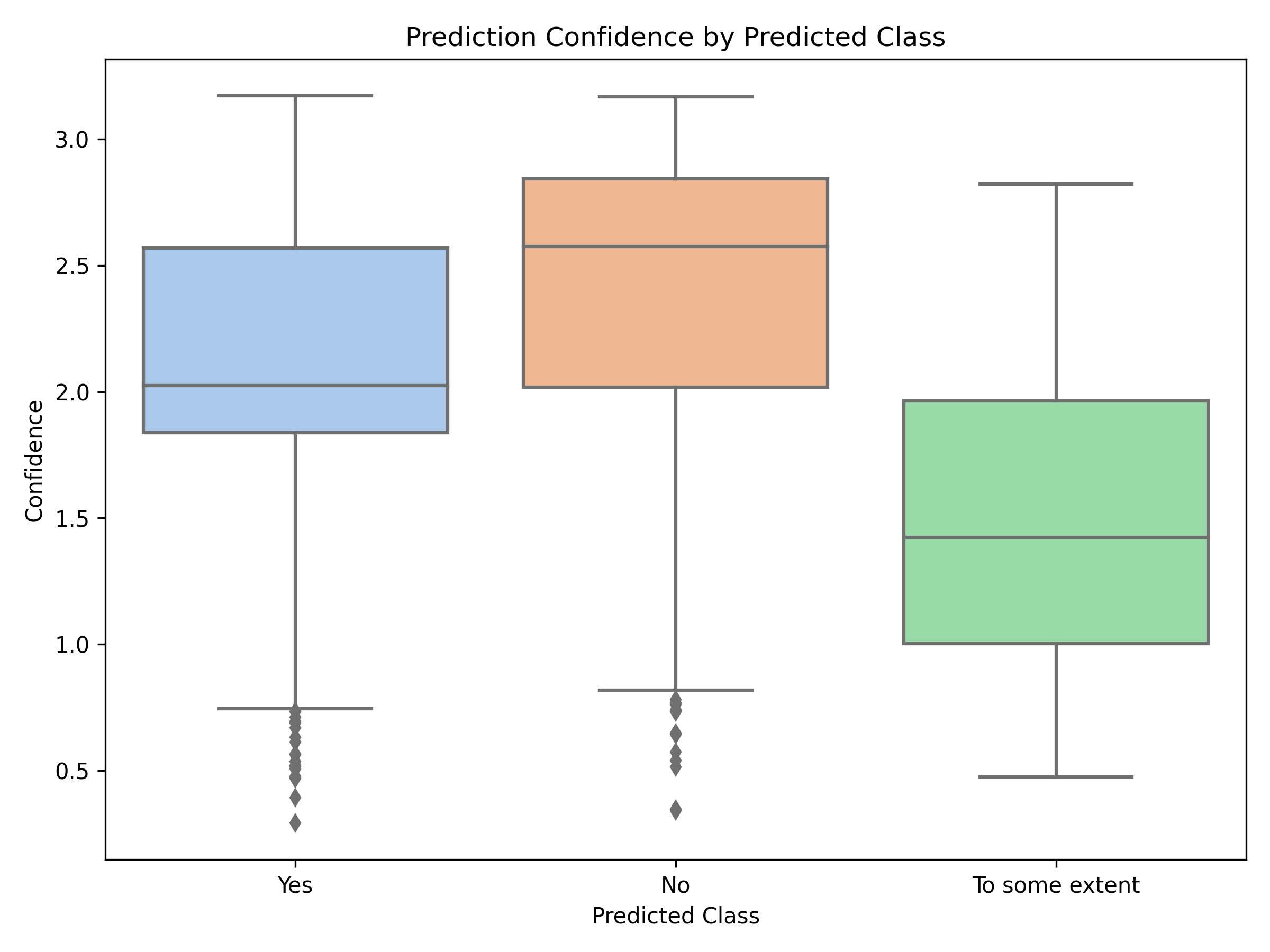}
\caption{Boxplot of confidence by predicted class (Track~2). “No” and “Yes” predictions show higher confidence than “To some extent.”}
\label{fig:task2_conf_box}
\end{figure}

In both tasks, the confidence distribution is skewed toward the middle range (1.5–3.0), with multiple local peaks. This suggests that while the model often makes moderately confident predictions, it does not frequently commit to extremely low or high confidence outputs. The spiked clusters in Track~2 (Figure~\ref{fig:task2_conf_hist}) hint at calibration artifacts possibly introduced by ensemble averaging. Despite ensemble smoothing, we still observe confidence saturation for some predictions near 3.5, particularly on easier instances.

To better understand class-specific behavior, we examined boxplots of prediction confidence grouped by predicted label (Figures~\ref{fig:task1_conf_box} and \ref{fig:task2_conf_box}). In both tasks, predictions labeled as “No” tend to have higher median confidence compared to “To some extent,” reflecting that the model is more certain when asserting a complete absence of error. Predictions for “To some extent” exhibit both lower median confidence and greater spread—supporting earlier findings that this category is harder to classify due to its inherent ambiguity. Interestingly, in Track~1, “Yes” predictions also show relatively high confidence, indicating that the model treats full error recognition as a more decisive signal than partial acknowledgment.

These confidence trends are broadly aligned with our confusion matrix analysis: “To some extent” is not only the most frequently confused class but also the one with the least confident predictions. This highlights a key challenge in pedagogical feedback modeling—the need to model uncertainty explicitly, especially in borderline cases. Future work could explore temperature scaling or Bayesian ensembling to better calibrate prediction confidence, particularly for interpretability in high-stakes educational settings.

\section{Limitations}

Despite the strong results achieved by our ensemble MPNet-based system, several limitations warrant discussion:

\textbf{Confidence Calibration}: Our ensemble's probability estimates are not well-calibrated. Overconfident predictions—such as assigning high confidence to incorrect labels—can be problematic in deployment, especially if these scores are used to trigger interventions. Due to time constraints, we did not apply calibration techniques like temperature scaling. Recent studies have introduced Adaptive Temperature Scaling (ATS), a post-hoc calibration method that predicts a temperature scaling parameter for each token prediction, improving calibration by over 10-50\% across various benchmarks \cite{ats2024}.

\textbf{Label Ambiguity}: The boundary between “Yes” and “To some extent” is inherently subjective. Some errors reflect annotation uncertainty rather than true model failure, placing an upper bound on performance. Modeling the task as ordinal or probabilistic, rather than categorical, could better capture this continuum. Ordinal classification approaches have been explored to address such challenges in tasks with naturally ordered labels \cite{ordinal2023}.

\textbf{Model Scope and Efficiency}: MPNet-base, while efficient, is not specialized for educational dialogue and may struggle with nuanced tutor-student interactions or long responses. Moreover, we trained separate models for each track. Employing a larger backbone, or a multitask setup could improve generalization. However, while effective, the ensemble adds computational overhead compared to a single fine-tuned large language model (LLM).

\section{Conclusion}

We presented Team BD’s system for the BEA 2025 Shared Task on Mistake Identification and Location in tutor responses. Our approach centered on fine-tuning MPNet, a pretrained Transformer model, and employing an ensemble of models to enhance robustness. By utilizing class-weighted loss and grouped cross-validation, we effectively addressed data imbalance and maximized the utility of the available training data. The resulting system achieved high accuracy and macro-F1 scores on both Track 1 and Track 2, demonstrating the viability of our approach for evaluating AI tutor responses.

Beyond quantitative results, we conducted extensive analyses to understand our model’s behavior. We observed that while the model excels at identifying clear-cut cases of tutor recognition or non-recognition of errors, it struggles with borderline cases involving partial acknowledgment. Visualization of the model’s embedding space revealed that these borderline cases are indeed less separable, reflecting an inherent challenge in the task. By categorizing our model’s common errors, we highlighted areas for improvement, such as better handling of implicit cues and calibrating model confidence.

In future work, we aim to explore multi-task learning across the different evaluation dimensions (e.g., training a single model to predict all four labels simultaneously). This approach could enable the model to leverage signals from one aspect (like providing guidance) to inform another (like mistake identification). We also consider incorporating larger language models or adapter-based approaches that can leverage the knowledge of LLMs—which were used to generate some tutor responses—to improve classification. Finally, techniques for model calibration \cite{chen2023close} and incorporating contextual knowledge (such as specific math domain concepts) could enhance the system's reliability and interpretability.

Overall, we believe our ensemble MPNet approach provides a strong baseline for these novel tasks. We hope that our detailed analysis of errors and the insights gained will inform future improvements in building AI tutors that can understand and respond to student mistakes with pedagogical sensitivity. Ultimately, the goal is to advance automated tutoring systems that not only generate helpful responses but also evaluate and refine their feedback in a pedagogically effective manner—the shared task is a step in this direction, and our work contributes to that ongoing journey.

\section*{Acknowledgments}
We thank the BEA organizers and annotators.  
Code and models will be released at: \url{https://github.com/ShadmanRohan/team-bd-bea25}

\bibliography{custom}

\appendix
\section*{Appendix}
\label{sec:appendix}

\paragraph{Implementation Details.} 
All models were trained on a single NVIDIA 3090 GPU. Each fold in cross-validation took approximately 25 minutes. We used the Hugging Face Transformers library for MPNet fine-tuning, with batch size 16, learning rate $2\mathrm{e}{-5}$, and early stopping based on validation macro-F1.




\begin{table*}[t]
  \centering
  \small
  \begin{tabular}{lccccc c}
    \toprule
    \textbf{Category}           & \textbf{Phi3} & \textbf{Mistral} & \textbf{Llama-3.1-8B} & \textbf{Llama-3.1-405B} & \textbf{GPT-4} & \textbf{Total} \\
    \midrule
    Extra Info                  & 1             & 0                & 1                   & 11                     & 1             & 14             \\
    Appended Dialogue Trimming        & 19            & 0                & 0                   & 0                      & 0             & 19             \\
    Code Abstraction                 & 2             & 0                & 0                   & 0                      & 0             & 2              \\
    Punctuation Cleanup           & 3             & 2                & 0                   & 0                      & 0             & 5              \\
    \midrule
    \textbf{Totals}             & \textbf{25}   & \textbf{2}       & \textbf{1}          & \textbf{11}            & \textbf{1}    & \textbf{40}    \\
    \bottomrule
  \end{tabular}
  \caption{Model‐specific frequencies of manual cleanup operations on tutor responses.}
  \label{tab:preprocessing-wide}
\end{table*}

\paragraph{Class Weights.} 
To mitigate class imbalance, we applied inverse frequency class weighting in the cross-entropy loss function:
\[
w_c = \frac{1}{\log(f_c + \epsilon)},
\]
where \( f_c \) is the frequency of class \( c \) and \( \epsilon = 1.05 \).

\paragraph{Hyperparameter Search.} 
We performed grid search over learning rates \{1e-5, 2e-5, 3e-5\} and batch sizes \{8, 16\}. The best configuration was selected based on average macro-F1 over the cross-validation folds.

\paragraph{Reproducibility.} 
We fixed all random seeds to 42 and set PyTorch to deterministic mode. Our code will be made publicly available upon publication.

\begin{table*}[t]
\centering
\small
\begin{tabular}{lcc}
\toprule
\textbf{Model} & \textbf{Mistake Identification} & \textbf{Mistake Location} \\
\midrule
BERT         & \textbf{0.8703} & \textbf{0.7025} \\
RoBERTa      & 0.7816          & 0.6551 \\
DeBERTa      & 0.8576          & \textbf{0.7025} \\
ELECTRA      & 0.8513          & 0.6266 \\
MPNet        & 0.8639          & 0.6203 \\
NeoBERT      & 0.8513          & 0.6677 \\
Logistic Regression & 0.7880    & 0.6139 \\
Random Forest       & 0.8260    & 0.6551 \\
Gradient Boosting   & 0.8418    & 0.6519 \\
SVM                 & 0.7785    & 0.6110 \\
LightGBM            & 0.8418    & 0.6551 \\
XGBoost             & 0.8386    & 0.6646 \\
CatBoost            & 0.8196    & 0.6582 \\
\bottomrule
\end{tabular}
\caption{Macro-F1 scores for Mistake Identification and Mistake Location tasks across Transformer models and TF-IDF + traditional classifiers. Best results per column are bolded.}
\label{tab:appendix-transposed}
\end{table*}

\end{document}